# MobileDenseNet: A new approach to object detection on mobile devices

Mohammad Hajizadeh[a] (mohammad.hajizadeh@gmail.com), Mohammad Sabokrou[b] (sabokro@gmail.com), Adel Rahmani[a] (rahmani@iust.ac.ir)

[a] School of Computer Engineering, Iran University of Science and Technology, Tehran, Iran

[b] School of Computer Science, IPM Institute For Research In Fundamental Sciences, Tehran, Iran

**Corresponding Author:**

Adel Rahmani

School of Computer Engineering, Iran University of Science and Technology, Tehran, Iran

Email: rahmani@iust.ac.ir

**Abstract**

Object detection problem solving has developed greatly within the past few years. There is a need for lighter models in instances where hardware limitations exist, as well as a demand for models to be tailored to mobile devices. In this article, we will assess the methods used when creating algorithms that address these issues. The main goal of this article is to increase accuracy in state-of-the-art algorithms while maintaining speed and real-time efficiency. The most significant issues in one-stage object detection pertains to small objects and inaccurate localization. As a solution, we created a new network by the name of "MobileDenseNet" suitable for embedded systems. We also developed a light neck "FCPNLite" for mobile devices that will aid with the detection of small objects. Our research revealed that very few papers cited necks in embedded systems. What differentiates our network from others' is our use of concatenation features. A small yet significant change to the head of the network amplified accuracy without increasing speed or limiting parameters. In short, our focus on the challenging CoCo and Pascal VOC datasets were 24.8 and 76.8 in percentage terms respectively - a rate higher than that recorded by other state-of-the-art systems thus far. Our network is able to increase accuracy while maintaining real-time efficiency on mobile devices. We calculated operational speed on Pixel 3 (Snapdragon 845) to 22.8 fps. The source code of this research is available on https://github.com/hajizadeh/MobileDenseNet.

## 1. Introduction

Neural networks have improved greatly in recent years - so much so that many tasks are either highly difficult or virtually impossible to achieve without them. In field of computer vision, these networks have been used in classification, localization, object detection, and segmentation. One major application of deep neural networks is object detection.

Various methods have been developed for one-stage object detection in recent years. One can cite SSD (Liu et al., 2016), YOLO (Redmon et al., 2016; Redmon & Farhadi, 2017, 2018), RetinaNet (Lin et al., 2017b), EfficientDet (Tan et al., 2020), etc. One persistent problem that all aforementioned examples struggle with is the application of these systems to situations where hardware limitations exist. The goal of our research is to focus on this issue specifically by developing networks applicable to real world needs on mobile CPUs operating in real time (Li et al., 2018; Qin et al., 2019; Wang et al., 2018; Tang et al., 2020; Xiong et al., 2021). In order to achieve this, we used a small backbone network (MobileNetV1) (Howard et al., 2017) and personalized its structure, ultimately creating "MobileDenseNet". We also developed necks "FCPNLite" and "SSDCLite". Our experience suggests a network with slight differences to "MobileNetV1" is suitable enough and can even perform better in object detec-
tion.

Our main goals were developing a lightweight algorithm with improved data flow transmission between its layers, while achieving real-time operation on mobile CPUs. We used depthwise separable convolution (Sifre & Mallat, 2014) and bottlenecks to decrease parameters and increase speed. We utilized concatenation in the place of "add" to boost the transfer of information in the neck and protect its features. We have realized that using "add" might lead to the loss of useful information. The details of such instances will be explained in full in section 3. Also, certain details for Subsampling in the MobileDenseNet network have changed compared to other MobileNets networks (Howard et al., 2017;

Sandler et al., 2018; Howard et al., 2019), which from our point of view and in our experiences, are effective in improving network accuracy. We will explain our results as follows.

An object detection network consists of 3 parts: Backbone, Neck and Head (Bochkovskiy et al., 2020). Our focus has been to optimize the first two components, as well as implement a small but useful change for the third component.

An overall overview for our proposed method has been shown in figure 1.

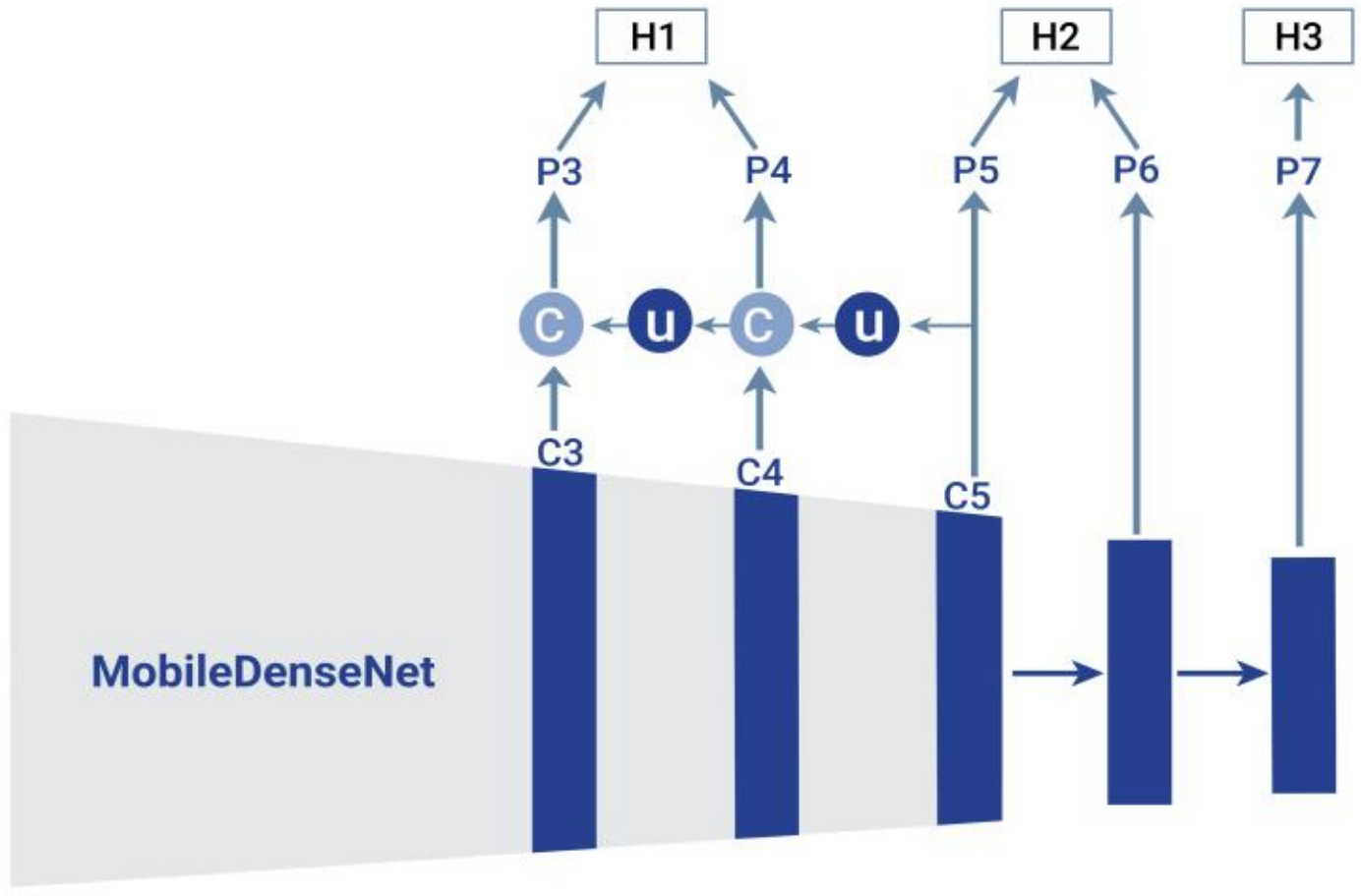

Figure 1: Overall proposed method for object detection

As will be explained, we combined our MobileDenseNet network with a number of Necks (Lin et al., 2017a; Fu et al., 2017; Tan et al., 2020) and witnessed results that surpassed those seen in state-of-the-art papers related to embedded object detection. No credible article had thus far succeeded in using Necks in embedded algorithms in real-time. How we ensured the Neck remained lite will be described in full in section 3. We also changed the Head and achieved different and better results compared to similar papers. The AP increased when the heads had separate weights compared to when all weights were shared. We also evaluated our model on the CoCo Challenge Dataset (Lin et al., 2014) and the Pascal VOC (Everingham et al., 2015). Our network achieved an accuracy of 24.8% AP on the CoCo database with AP evaluation criteria and 76.8% mAP on the Pascal VOC. It is worth noting that IoU 0.95 is not required for general applications (Redmon & Farhadi, 2018). When even the human eye is not able to recognize objects within this range, why focus on the issue? It can be said that accuracy within the old standard, AP50, is as important as the main criterion. In this scenario, we reached an accuracy of 43.5% AP50 on the COCO dataset.

Certain other details that are often universal for all object recognition problems such as the size of anchor boxes (Ren et al., 2015), the use of various loss functions, and data augmentation (Howard, 2013) have changed slightly compared to popular object detection methods; They are more similar to SSD (Liu et al., 2016) which will be explained in its own subsection. In section 2, we will give an overview of the work done in this field. section 3 examines in detail our proposed methods and model. section 4 also describes our evaluation and experiences in this field. section 5 is devoted to the final summary and conclusion.

## 2. Related Work

### *2.1. Generic Object Detection*

Object detection is the task of locating object instances in natural images and classify their categories. Many methods have been proposed for this task in recent years (Liu et al., 2020).

The most popular one-stage methods are Yolo (Redmon et al., 2016), SSD (Liu et al., 2016), RetinaNet (Lin et al., 2017b) and EfficientDet (Tan et al., 2020). When we refer to Yolo, we mean YoloV3 (Redmon & Farhadi, 2018), which is suggested by the authors of this paper to be similar to our use of YoloTiny (Adarsh et al., 2020), and uses a much smaller model; However, this model is not suitable for applications with hardware limitations. The original SSD also used VGG (Simonyan & Zisserman, 2014), which is still not suitable for hardware limitations; but one can make use of the SSD design and change the backbone or feature extractor to use the SSD for hardware limitations. Other methods are mentioned that although are not suitable for this application, one can draw ideas from.

RetinaNet and EfficientDet articles also use backbones that are not suitable for embedded applications. The mathematical operation rate of the smallest EfficientDet network (Tan & Le, 2019) combined with Neck and Head is 2.5B operation, which may be appropriate for hardware constraints; but it is still heavier than other existing networks. Other EffcientDets from B1 to B7 and networks like ResNet50 and ResNet101 (He et al., 2016) are not suitable solutions to the problem in any way.

The YoloV4 (Bochkovskiy et al., 2020) paper contains useful terms for the topic. By definition, if you add a function to your method that does not increase the execution time and volume of your model but increases accuracy, it is called a "bag of freebies". Examples of this are changing the loss function and data augmentation. Alternatively, changes that make the model heavier and increase accuracy are called "bag of specials"- similar to adding feature pyramid networks (Lin et al., 2017a) or skip connections (Liu et al., 2018) in various forms. These items have less use in an embedded context, because the main goal is to prevent the increase in size and the number of mathematical operations.

### *2.2. Efficient (light) Object detection*

The use of depthwise separable convolution (Sifre & Mallat, 2014) instead of standard convolution revolutionized hardware constraint applications. One of the main suggestions for backbone or feature extractors was MobilenetV1 and MobilenetV2 networks (Howard et al., 2017; Sandler et al., 2018). The main difference between these networks and other previous networks was the use of depthwise separable convolutions, which reduces the number of parameters and calculations by nearly 8.9 times. SSDLite (Sandler et al., 2018) is also proposed by MobilenetV2 as an alternative to the SSD network for object detection and is mounted on the backbone. With these new descriptions, one can refer to the SSDLite as a Neck and Head.

The MobileDets article (Xiong et al., 2021) also improved accuracy by designing and proposing changes to the depthwise convolutions and inverted bottleneck layers proposed in MobileNetV2. By proposing fused inverted bottleneck layers and tucker convolution layers, this paper has been able to increase the accuracy on mobile CPUs for MobilenetV2 and MobileNetV3 by 1.9% and 1.7% AP respectively- without slowing down the execution of the CoCo datasets. In another project, Pelee (Wang et al., 2018), a model was designed that claimed to reduce size by up to 34% without compromising accuracy. The backbone they provide is 1.8 times faster than MobilnetV1 and MobilenetV2. Of course, the size of their final model, full object detection, is equal to 5.98 million parameters, which is still more than Mobilenet + SSD and MobilenetV2 + SSDLite; however, they claim that their accuracy on the CoCo challenge database has increased by 0.3 AP, which is not significant.

The same issue was addressed in a paper called Tiny-DSOD (Li et al., 2018) the following year. This article proposed a new block for the network called the depthwise dense block to address hardware limitations. A D-FPN was also used as a neckline in the network. The main advantage was that after PeleeNet, it was able to achieve a 23.2 AP resolution on the CoCo data by reducing the model size to 1.15M. Although ultimately, it has been successful in reducing network size and reducing the number of mathematical operations. ThunderNet (Qin et al., 2019) is one of the first real-time methods in the world of two-stage object detections. The paper manages to reach an accuracy of 19.2% at 24.1 fps on the CoCo database on the Mobile CPU. This article has reached real-time on GPU and Xeon CPU with an accuracy of 28.1% at 15.3 fps. This article has designed 3 different networks, SNet49, SNet146, SNet535, each with different accuracy and speed. Since the purpose of this article is to operate in real-time on mobile CPUs, we will compare our method with SNet49 which it's computational speed in more than 20 fps. We will disclose our proposed method to deal with this issue as follows.

## 3. Proposed Method

As a reminder and a start to this section, it should be mentioned that there are a series of factors that must be specified for object detection. Most articles do not address this issue well. All the main differences between the one-step Object Detection methods are in the following headings. The standard and basic factors of all the methods mentioned in section 2 are the same. These are as follows:

- Backbone
- Neck
- Head
- The number and size of feature maps
- Determining the area, scales, aspect ratios and number of anchor boxes
- Data augmentation and all other standard regularizers
- Network training such as optimizer, learning rate, learning rate scheduler, weight decay, batch size. In general, the answer to how to teach your final network.
- Use different loss functions for classification and localization

The first three parts produce our core network. In this section, the necessary changes can be applied based on different applications in order to achieve better accuracy and speed in both general and specific applications. The next two items determine the number of our anchor boxes. The next three items help us increase our accuracy without increasing or decreasing network size, speed, or anything else. The main focus of our article is on items 1, 2, 3, 4 and 5. Small changes have also been made in loss function. The rest is selected from those presented in other rticles. Also, the recommended network input size for object detection is 320x320, which, similar to others, has been tested for evaluation.

The various options were reviewed and the best of them selected. Larger sizes have not been tested due to increased FLOP and inefficiency for our intended use.

*3.1. Backbone*

The core of the model, in our opinion, is the backbone. The main network generates features for us so that other parts can use these features to extract objects. Networks used in various object detections include VGG, Resnet, Darknet, and EfficientNet. These networks are not suitable for real-world applications with hardware limitations. Some articles use MobileNet and MobileNetV2 (Howard et al., 2017; Sandler et al., 2018). In our opinion, and based on our experience and experiments, these networks are not suitable for object detection and especially Localization. The reason for their inadequacy is that these networks were originally designed for classification and do not have enough lowlevel information flow features. An article also used PleeNet, which they claim has 66% of MobileNet parameters; But it is still not suitable for localization and its accuracy has not increased.

For localization and good object placement accuracy, we need a network that also has low level features. We want to design a network where the flow of information is suitable for localization and also with real hardware such as that in the CPU of a mobile device that needs to run in real-time. As a result, we designed and implemented the MobileDenseNet network.

As mentioned, Mobilenets use depthwise separable convolution instead of convolution. There is no skip connection in MobilenetV1 network, and so it is not suitable for localization. In MobilenetV2, skip connections are like Resnet and are accomplished with add. It can be said to some extent that inverted residual connections + Mobilenet = MobilenetV2. In our opinion, if we use concatenation instead of add, we will allow the network to select from those features with a 1x1 bottleneck convolution as a feature selector. Also, using add generates new features that differ from the origin. Finally, according to our experience, if we use concatenation instead of add, the properties of the lower layers reach our object detection more satisfactorily and results in better accuracy. Of course, when we decide to use concatenation, it is clear that the number of parameters has increased. Therefore, it is necessary to use bottlenecks. It should be noted that if the number of dense connections increases past one point, they greatly increase the computational cost of the network so we must avoid their overuse. DenseNet (Huang et al., 2017), for example, has used a large number of skip connections as concatenation, which has severely slowed down the network on CPU.

Another problem with one-stage methods is that they are not equipped to detect small objects. We have largely solved this problem by making changes to the proposed network. Since all backbones are provided from the beginning for image classification problem, an important problem in all referenced backbones is that the number of layers in second and third blocks are very low. And because we need better information from these blocks that we use them for C3 and C4 output from middle of the backbone, we decided to slightly increase the number of layers in the second and third blocks of the network so that the feature maps taken from these layers have better and more comprehensive features. The result will be described in section Four.

In short, one can say:

Mobilenet (Enhanced version) + Dense connection (Concatenation) + Bottleneck = MobileDenseNet

Our network, like other backbone networks, contains 5 main blocks. Similar to Mobilenet, we have considered the necessary parameters to determine the final number of features in each block. It should also be noted that different architectures can be used for different applications. The MobileDenseNet network has the flexibility to increase or decrease the number of layers per block for different applications. Our core network architecture consists of a number of depthwise separable convolution blocks. The only major difference is that in some cases dense connections are used. Our proposed architecture for object detection has been shown in the figure 2.

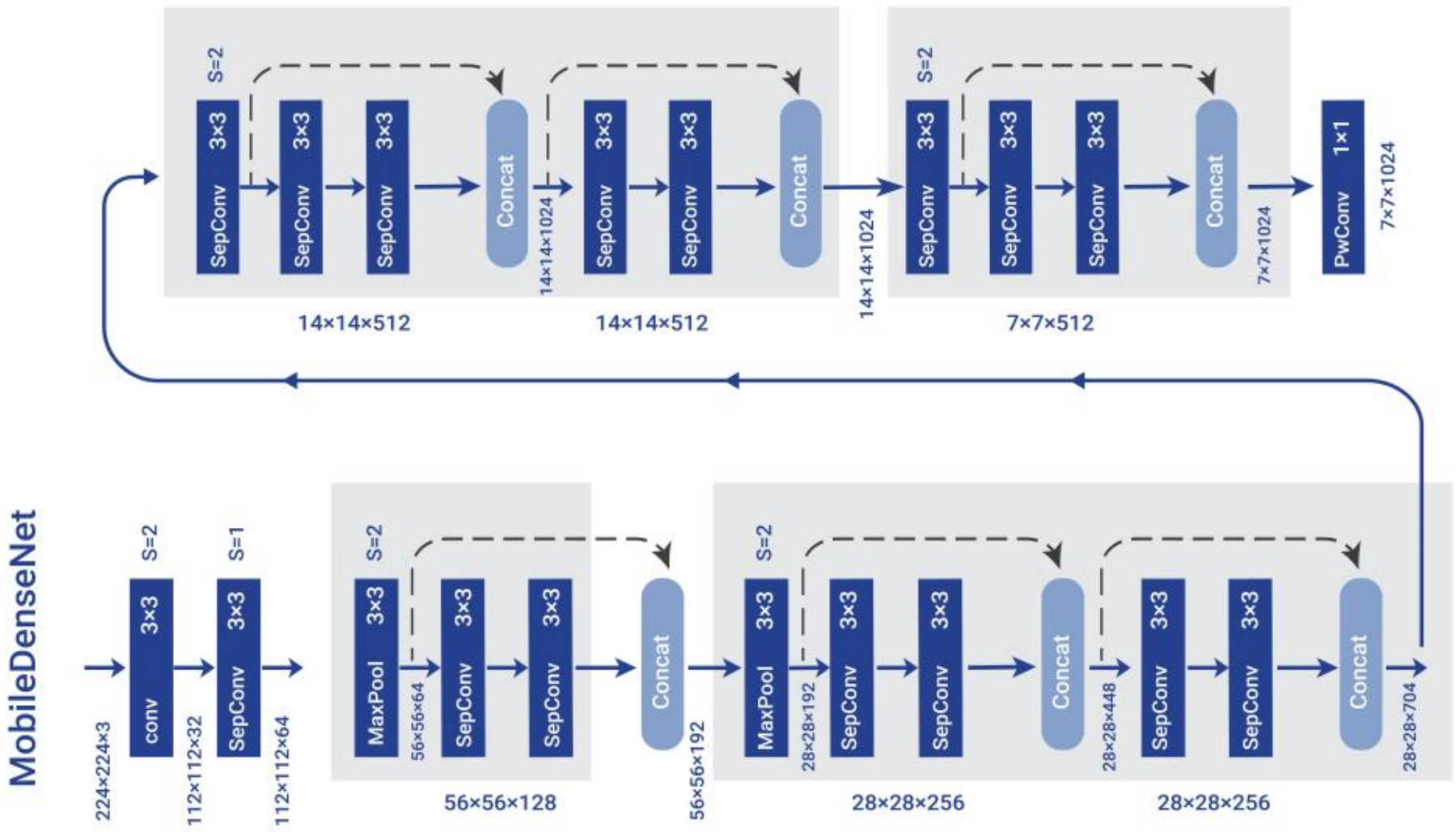

Figure 2: MobileDenseNet backbone network architecture

We need to have 3 outputs from our MobileDenseNet backbone. The formula can be defined as

$$C3, C4, C5 = MobileDenseNet(InputImage)$$

Where $C3$, $C4$ and $C5$ are the outputs of third, fourth and fifth block, respectively. Also we have defined two new outputs from $C5$ as

$$C6 = SeparableDepthwiseConvolution(C5, stride = 2)$$

$$C7 = SeparableDepthwiseConvolution(C6, stride = 2)$$

*3.2. Neck*

As our second point, Neck, many changes have been implemented. FPN, Bi-FPN, NAS-FPN and CPAM (Tan et al., 2020; Ghiasi et al., 2019; Liu et al., 2018) are not suitable for embedded applications. Many of these, like the model used in EfficientDet (Tan et al., 2020), are too non-specific and come with heavy hardware limitations. They bring a large number of parameters and computational cost to the network.

We also proposed different necks with the goal to create a result more efficient than SSDLite. The main problem with single-stage object detection networks is their low accuracy when it comes to small objects. In general experiences, adding and duplicating these pyramids as other papers have done is time consuming with small increases in accuracy. One stream of information is enough to combine low level features and high-level features. The reason for using concatenation, is to allow the network itself to use any of the features it wants. The add function creates new features, and the concatenation allows low-level features to reach the final stage more untouched. We will look at some of the suggested necks in the evaluation section. The final suggested Neck can be viewed in the figure below. In this neck, instead of a large number of connections, we have only used a limited number of connections to improve the accuracy of small objects.

For the Neck part, we converted the first Neck known as FPN (Lin et al., 2017a) with a few tweaks to add concatenation instead of add, use a bottleneck to reduce computations, and use depthwise separable convolution instead of the usual convolution, which has a large number of parameters. We created our own proposal called FCPNLite, feature concatenation pyramids network lite (Fig. 3 center) that can be formulated as

$$P3 = PointwiseConvolution(Concat(C3, Upsample(C4)))$$

$$P4 = PointwiseConvolution(Concat(C4, Upsample(C5)))$$

$$P5 = PointwiseConvolution(Concat(C5, Upsample(C6)))$$
$$P6 = PointwiseConvolution(Concat(C6, Upsample(C7)))$$
$$P7 = C7$$

Where $C3$, $C4$, $C5$, $C6$ and $C7$ are the outputs from backbone and $P3$, $P4$, $P5$, $P6$ and $P7$ are the outputs from neck.

Also, if Add is used like the original article, a new Neck called FPNLite will be created (Fig. 3 left). After the experiments, we came to the conclusion that not all feature maps need to be connected with concatenation or add. As mentioned, the overuse of concatenation and add significantly increases the amount of FLOPs and computational cost. Network accuracy can be increased with a limited number of connections. Since the features for C5, C6, and C7 are similar in strength, there is no need for up-sampling and concatenating. Ultimately, a simpler Neck than FCPNLite that was highly accurate with a lower computational cost. We named this Neck SSDCLite, SSD concatenation lite (Fig. 3 right). The main difference is that the number of connections in this neck is limited.

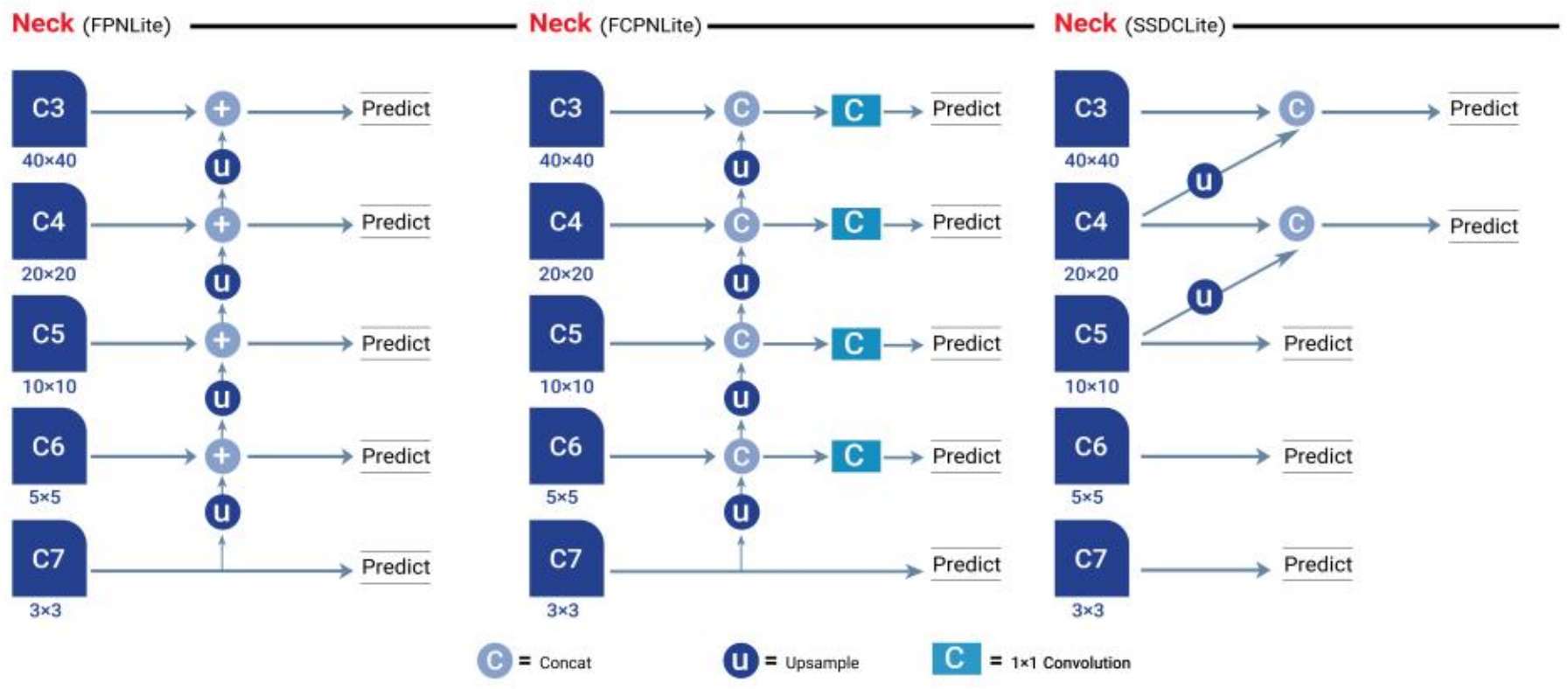

Figure 3: Different neck architectures. FPNLite (left), FCPNLite (center), SSDCLite (right)

*3.3. Head*

For the Head, as other networks, we have used a number of classification and box headers; with the difference that their weight is not entirely share. We used our novel idea, half-share. This means that if the weight of the whole head is shared with each other (Lin et al., 2017b,a), it will be difficult for the network to extract very small and very large objects using a shared weight. Since it can be said that the dimensions and shapes of the objects in both feature maps are quite close, we shared the weight of every two heads in one row. We shared the weights of two consecutive heads, increasing the AP to 1.1% when the weights are common (Lin et al., 2017b,a) and 0.3% when the weights are not common (Liu et al., 2016). All 5 feature maps from p3 to p7 are used for size 256 filters, we have shared weights p3 with p4, p5 with p6 and p7 with 3x3 size is separated from others. We have examined the different modes of share or no share between different heads (Fig. 4), which will be evaluated in the next section. Our final choice is the Half-Share model. Our Head part can be formulated as

$$predictions(40 * 40 * 9) = H1(P3)$$

$$predictions(20 * 20 * 9) = H1(P4)$$

$$predictions(10 * 10 * 9) = H2(P5)$$

$$predictions(5 * 5 * 9) = H2(P6) \; predictions(3 *$$

$$3 * 9) = H3(P7)$$

Where $P3$, $P4$, $P5$, $P6$ and $P7$ are the outputs from neck and $H1$, $H2$ and $H3$ are three different head modules with separate weights.

## 4. Experimental Results

This section will evaluate and review the final proposed model. Each of the suggested sections of this article for object detection will be considered in this section. We will begin with the head since it analysis is more simple. The description of each part is given in full in section 3.

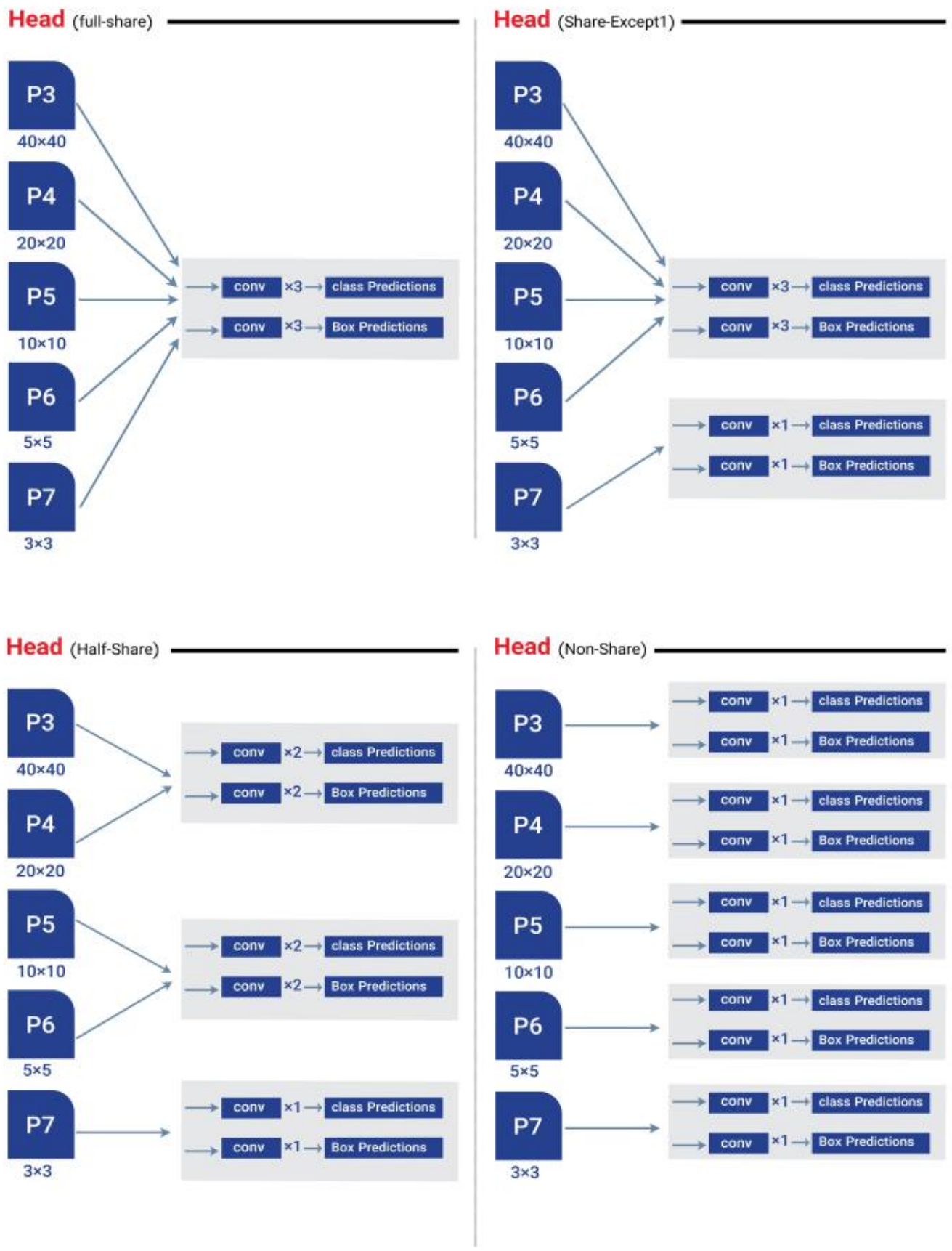

Figure 4: Different head architectures. Full-share (left-top), half-share (leftdown), share-except1 (right-top), non-share (right-down)

### *4.1. Dataset*

COCO is a large-scale object detection, segmentation, and captioning dataset. It consist of 80 object categories in more than 330K images and 1.5 million object instances (Lin et al., 2014).

Pascal VOC is another dataset for object detection. It consists of 20 categories in 11.5K training and validation images and more than 27K ROI annotated objects (Everingham et al., 2015).

*4.2. Anchor or Prior Boxes*

In our network, 5 feature maps have been used. These areas starts from $24^2$ to $384^2$ on $P3$ to $P7$. Due to the lite nature of the network, adding a feature map does not add much to our computational cost and can be used. A number of anchor boxes are provided in sizes $2^0$ and $2^{1/2}$ times the size of the feature map area. These sizes are determined by the size of each grid and their respective field. For example, a feature map with dimensions of 10x10 and input size of 320x320. Each of the grids represents a 32x32 part of the original image. So, the dimensions of the anchor boxes in this feature map are equal to 96 and 135.

There are also different dimensions for each of these sizes. Our dimensions are 1:1, 1:2, 1:3, 2:1 and 3:1.

We have suggested a criterion when determining the best box. This criterion includes calculating the percentage of ground truth boxes for which a suitable anchor box cannot be found. In cases such as 1:4, 4:1 the error was greater. We tested the different modes between these choices in terms of size and dimensions, and the best choice was 1:1, 1:2, 1:3, 2:1 and 3:1. Taking this into account, we have 10 anchor boxes for each grid in each feature map. Our proposed anchor boxes, pushed our AP by 0.4 percent. Obviously, the higher the number of anchor boxes, the more likely it is that the accuracy will increase; but the computational cost will also be relatively high. Because the computational cost is important to our paper, we can't consider employing 100,000 anchor boxes as RetinaNet does. In our work, standard anchor box encoding/decoding is used. (Liu et al., 2016; Ren et al., 2015). Some examples of different feature map sizes and their receptive field maps on input has been shown in figure 5.

*4.3. Loss Function*

In each object detection project, two separate loss functions must be selected for classification and localization. We implemented three different loss function and anchor box assignment proposed in SSD, RetinaNet and YoloV3.

For classification, the best choices are to use either cross entropy with hard negative mining or focal loss. In our experiments, cross entropy with hard negative mining (Liu et al., 2016; Ren et al., 2015) achieved better accuracy. Using focal loss slightly reduced accuracy. A small change from the original SSD loss function was that we increased the positive/negative ratio to 1:6 and changed the Negative class coefficient from 1.0 to 0.5. Both negative and positive classes normalized by number of assigned prior boxes. The final loss function for classification can be defined as

$$CE = (y * log(p)) + (0.5 * (1 - y) * log(1 - p))$$

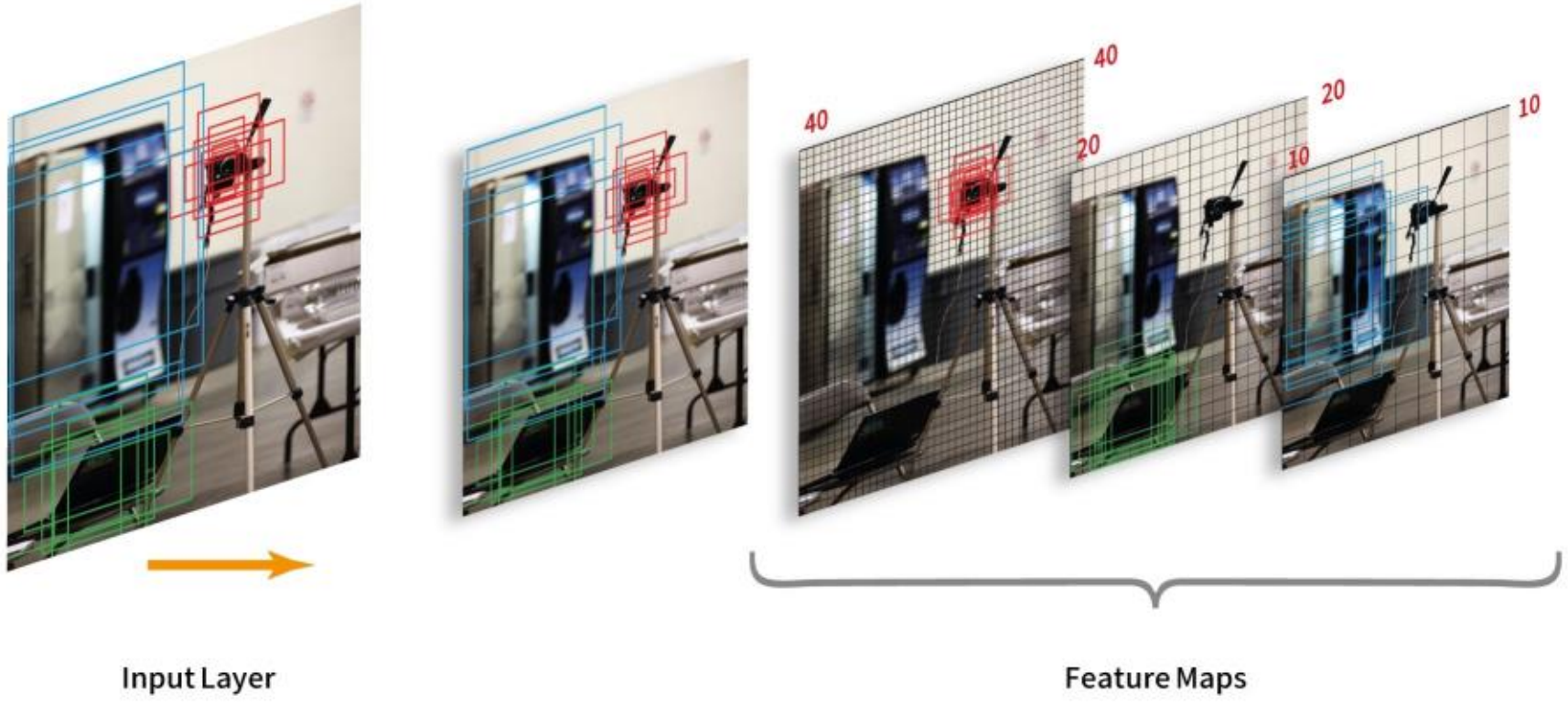

Figure 5: Examples of different feature map sizes and their receptive field maps

Where $p$ is our prediction and $y$ is the ground truth.

For localization, SSD and Yolo articles use L1 smooth loss (Liu et al., 2016; Redmon et al., 2016). The main problem with these functions was that each of the four variables was minimized separately and did not affect each other.

Newer articles unveiled IoU Loss (Zheng et al., 2020).

$$IoU = (B \cap B^{gt})/(B \cup B^{gt})$$

$$Loss_{IoU} = 1 - (B \cap B^{gt})/(B \cup B^{gt})$$

Where $B$ is the predicted box and $B^{gt}$ is the ground truth.

New versions of this loss function were introduced to fix the convergence issue of the function and improve upon it. We found that DIoU loss performed better than others (Zheng et al., 2020). The final loss function for localization can be defined as

$$DIoU = 1 - IoU + (p^2(b,b^{gt}))/c^2$$

Where $b$ and $b^{gt}$ denote the central points of $B$ and $B^{gt}$, $p(.)$ is the Euclidean distance, and $c$ is the diagonal length of the smallest enclosing box covering the two boxes (Zheng et al., 2020).

The final target function in out setting is

$$Loss = (CE + a * DIoU)/N$$

Where $N$ is number of positive anchor boxes and $a$ is a coefficient for balancing localization and classification losses. In our setting $a$ is equal to 1.

*4.4. Data Augmentation and Other Regularizers*

As mentioned, the advantage of these options is that they are solely used during training and have no negative impact on execution time. YoloV4 article mentions them as a “bag of freebies” (Bochkovskiy et al., 2020). The positive impact of these cases on the accuracy is less than the network; but there is no cost to use them as far as runtime in concerned.

In our proposed method, after each depthwise convolution and pointwise convolution layer, we have used the proposed batch normalization layer (Yun et al.,2019). ReLU6 (Howard et al., 2017) was used after each batch normalization. For augmentation, standard color data augmentation methods have been used (Howard, 2013). Also, with a probability of 0.5, the photos are flipped horizontally and some parts of the photo are cropped out and zoom in randomly.

*4.5. How to Train our Network?*

We first trained our backbone network MobileDenseNet on ImageNet 2012 with 1K classes (Russakovsky et al., 2015). This training started with a learning rate of 0.01 and was multiplied by 0.1 after 120 thousand iterations and multiplied again by 0.1 after 160 thousand iterations. For this part, weight decay with a value of 0.002 and momentum equal to 0.9 were used. The batch size was considered equal to 256. We have done our training on 8 V100 GPUs.

Using our backbone network and mounting the suggested Neck and Head, we trained the final network for object detection end-to-end using SGD with an initial learning rate of 0.001. After 40,000 iterations, this value changed to 0.0001, and after 60,000 iterations, this value equaled to 0.00001. Our training continued at 80,000 iterations. We used weight decay equal to 0.0005 and momentum equal to 0.9. The batch size value at this stage was 32.

As mentioned in ThunderNet (Qin et al., 2019) and LightDet (Tang et al., 2020), final networks have trained with multi-scale training. In this article, we were able to get better results in our final model by applying multi-scale training method in our process.

*4.6. Head Analysis*

Table 1 shows that if we are in half-share mode, we will reach the maximum accuracy. These experiments were executed by holding the rest of the cases constant, namely the backbone and Neck.

| Network With Different Heads | FLOPs (B) | Params (M) | AP |
|---|---|---|---|
| MobileNetV1 + SSDLite + Full-Share | 1.57 | 4.7 | 20.9 |
| MobileNetV1 + SSDLite + Share-Except1 | 1.57 | 4.8 | 21.9 |
| MobileNetV1 + SSDLite + Half-Share | 1.36 | 5.1 | 22.2 |
| MobileNetV1 + SSDLite + Non-Share | 1.21 | 5.0 | 21.8 |
| MobileDenseNet + SSDLite + Full-Share | 1.67 | 4.5 | 21.9 |
| MobileDenseNet + SSDLite + Share-Except1 | 1.66 | 4.5 | 22.7 |
| MobileDenseNet + SSDLite + Half-Share | 1.48 | 4.8 | **23.2** |
| MobileDenseNet + SSDLite + Non-Share | 1.29 | 4.6 | 22.9 |

Table 1: Different head analysis reported on the COCO dataset

This is due to the fact that objects that are close in size are better able to achieve greater generalization by sharing weights. Additionally, AP is greatly reduced if we have only one head whose total weight is shared between 5 feature maps. Of course, this is because the 3x3 layer is not very compatible with the rest. We evaluated the case where p3 to p6 are common and p7 is still separate.
The best-case scenario is when weights are shared 2 by 2.

*4.7. Neck Analysis*

As described in section 3, our proposed necks were implemented and tested, which can be seen in table 2 alongside their accuracy and speed. All Necks were compared with the Neck suggested in the MobilenetV2 article, SSDLite. FCPNLite and SSDCLite were the best. These show the power of dense connections and Necks. In addition, the main problem with FCPNLite is that it has slightly increased the computational cost. Therefore, SSDCLite is more efficient than SSDlite in terms of accuracy while increasing the computing cost by a very small margin.

| Network With Different Necks | FLOPs (B) | Params (M) | AP |
|---|---|---|---|
| MobileNetV1 + SSDLite (Howard et al., 2017) | 1.37 | 5.1 | 22.1 |
| MobileNetV1 + SSDCLite | 1.42 | 5.4 | 22.7 |
| MobileNetV1 + FPNLite | 1.47 | 5.4 | 20.2 |
| MobileNetV1 + FCPNLite | 1.65 | 5.8 | 23.1 |
| MobileDenseNet + SSDLite | 1.48 | 4.8 | 23.2 |
| MobileDenseNet + SSDCLite | 1.51 | 5.6 | 23.8 |
| MobileDenseNet + FPNLite | 1.50 | 5.6 | 21.8 |
| MobileDenseNet + FCPNLite | 1.64 | 5.8 | **24.2** |

Table 2: Different neck analysis reported on the COCO dataset

*4.8. Backbone Analysis*

As explained in previous sections, we have largely solved small object detection problem by making changes to the proposed network. Since all backbones are provided with low number of layers in second and third blocks. And because we need better information from these blocks that we use them for C3 and C4 output from middle of the backbone, we decided to slightly increase the number of layers in the second and third blocks of the network so that the feature maps taken from these layers have better and more comprehensive features.

| Network With Different backbones | FLOPs (B) | Params (M) | APs | APm | APl |
|---|---|---|---|---|---|
| MobileNetV1 + SSDLite (Howard et al., 2017) | 1.37 | 5.1 | 5.1 | 22.6 | 31.3 |
| MobileNetV2 + SSDLite | 0.8 | 4.3 | 5.0 | 22.3 | 31.5 |
| MobileDenseNet + SSDLite | 1.48 | 4.8 | 7.4 | 23.7 | 34.8 |
| MobileDenseNet + SSDCLite | 1.51 | 5.6 | 8.3 | 24.8 | 34.8 |
| MobileDenseNet + FCPNLite | 1.64 | 5.8 | 8.7 | 25.3 | 34.9 |

Table 3: Different backbone analysis reported on the COCO dataset.

APs, APm and APl stand for small, medium and large objects respectively. Small objects are objects with an area less than $32^2$ pixels, while medium objects are objects with an area between $32^2$ pixels and $96^2$ pixels and large objects are objects with an area more than $96^2$ pixels. In table 3, AP of small object shows significant improvements from mobilenet backbones. Due to the increased accuracy of medium and large APs, the overall result of object detection has also increased which will be explained in the next section.

*4.9. Final Method Analysis*

As explained, with changes in the MobileNetV1 network and the addition of a number of blocks and dense connections, as well as a change in subsampling and other similar details, we were able to increase the accuracy without increasing the size of the backbone network. Now you can see the final accuracy of our model.

AP50 is the average precision on IoU threshold equal to 0.5. AP75 is the average precision on IoU threshold equal to 0.75. AP is the mean of average precision on IoU=0.5 up to 0.95

| Full Network | input | FLOPs (B) | Params (M) | AP | AP50 | AP75 |
|---|---|---|---|---|---|---|
| YOLOv2 (Redmon et al., 2016) | 416 x 416 | 17.5 | 67.43 | 21.6 | 44.0 | 19.2 |
| YOLOv3 (Redmon & Farhadi, 2018) | 320 x 320 | – | 62.3 | – | 51.5 | – |
| SSD300 (Liu et al., 2016) | 300 x 300 | 34.36 | 34.3 | 25.1 | 43.1 | 25.8 |
| DSSD321 (Fu et al., 2017) | 321 x 321 | 22.3 | – | 28 | 46.1 | 29.2 |
| YOLOv3-Tiny (Adarsh et al., 2020) | 416 x 416 | – | 12.3 | – | 33.1 | |
| MobileNetV1 + SSDLite (Howard et al., 2017; Sandler et al., 2018) | 320 x 320 | 1.3 | 5.1 | 22.2 | – | – |
| MobileNetV2 + SSDLite (Sandler et al., 2018) | 320 x 320 | 0.8 | 4.3 | 22.1 | – | – |
| PeleeNet (Wang et al., 2018) | 304 x 304 | 1.29 | 5.98 | 22.4 | 38.3 | 22.9 |
| Tiny-DSOD (Li et al., 2018) | 300 x 300 | 1.12 | 1.15 | 23.2 | 40.4 | 22.8 |
| ThunderNet (SNet49) (Qin et al., 2019) | 320 x 320 | 0.26 | 4.8 | 19.2 | 33.7 | 19.7 |
| LightDet (Tang et al., 2020) | 320 x 320 | 0.50 | – | 24.0 | 42.7 | 24.5 |
| MobileDenseNet + SSDLite + Half-Share (Ours) | 320 x 320 | 1.48 | 4.8 | 23.2 | 38.2 | 23.7 |
| MobileDenseNet + SSDCLite + Half-Share (Ours) | 320 x 320 | 1.51 | 5.6 | 23.9 | 41.1 | 24.1 |
| MobileDenseNet + FCPNLite + Half-Share (Ours) | 320 x 320 | 1.64 | 5.8 | 24.2 | 42.6 | 24.5 |
| MobileDenseNet + FCPNLite + Half-Share (Ours) ms-train | 320 x 320 | 1.64 | 5.8 | **24.8** | 43.5 | 25.1 |

Table 4: Full method analysis reported on the COCO dataset. ms-train stands for a method that has been trained by multi scale training

In table 4, the final accuracy of our system, MobileDenseNet + FCPNLite + Half Share Head with multi scale training, can be seen on the COCO dataset. Our method was able to achieve a 2.6% increase in AP accuracy over MobilenetV1, 0.8% higher than LightDet, its competitor. We calculated operational speed on Pixel 3 (Snapdragon 845) to 22.8 fps. The main reason for this advantage is the proper use of Dense Connections and better weight sharing in the Head. We also found that increasing the number of skip connections and deepening the model in embedded applications is not efficient and can even increase the computational cost.

| Full Network | input | FLOPs (B) | Params (M) | mAP |
|---|---|---|---|---|
| YOLOv2 (Redmon et al., 2016) | 416 x 416 | 17.5 | 67.43 | 76.8 |
| SSD300 (Liu et al., 2016) | 300 x 300 | 34.36 | 34.3 | 77.2 |
| DSSD321 (Fu et al., 2017) | 321 x 321 | 22.3 | – | 78.6 |
| YOLOv3-Tiny (Adarsh et al., 2020) | 416 x 416 | – | 12.3 | 57.1 |
| MobileNetV1 + SSDLite (Howard et al., 2017; Sandler et al., 2018) | 320 x 320 | 1.3 | 5.1 | 68.0 |
| PeleeNet (Wang et al., 2018) | 304 x 304 | 1.29 | 5.98 | 70.9 |
| Tiny-DSOD (Li et al., 2018) | 300 x 300 | 1.12 | 1.15 | 72.1 |
| ThunderNet (SNet49) ms-train (Qin et al., 2019) | 320 x 320 | 0.26 | 4.8 | 70.1 |
| LightDet (Tang et al., 2020) | 320 x 320 | 0.50 | – | 74.0 |
| LightDet ms-train (Tang et al., 2020) | 320 x 320 | 0.50 | – | 75.5 |
| MobileDenseNet + SSDLite + Half-Share (Ours) | 320 x 320 | 1.48 | 4.8 | 73.9 |
| MobileDenseNet + SSDCLite + Half-Share (Ours) | 320 x 320 | 1.51 | 5.6 | 75.0 |
| MobileDenseNet + FCPNLite + Half-Share (Ours) | 320 x 320 | 1.64 | 5.8 | 75.6 |
| MobileDenseNet + FCPNLite + Half-Share (Ours) ms-train | 320 x 320 | 1.64 | 5.8 | **76.8** |

Table 5: Full method analysis reported on the Pascal VOC dataset. ms-train stands for a method that has been trained by multi scale training

In table 5, the overall accuracy of our method, MobileDenseNet + FCPNLite + Half Share Head with multi scale training, can be seen on the Pascal VOC dataset. Our method was able to achieve a 1.3% increase in mAP accuracy over LightDet.

ThunderNet (Qin et al., 2019) is the only two-stage method in our comparison. The reason is that these methods have good results but are not suitable for the real world scenarios. Two-stage methods are always better in terms of accuracy/speed trade-off, but in real-world problems (such as automatic driving) when the number objects in the scene increases, the amount of flops and computational cost increases accordingly, which is not suitable for real-time embedded devices. Meanwhile, one-stage methods like our proposed method are not sensitive to this variable.

Our goal in this paper was to maximize the use of hardware resources to improve model accuracy and real-time processing. Therefore, we focus on accuracy/speed trade-off. But in LightDet (Tang et al., 2020), the processing speed is reported on a NVIDIA 1080 GPU, and it also uses 17 Shuffle Blocks in its network, each of which has a Concatenation layer. Based on our experience, it greatly increases the inference time on embedded CPUs. For this reason, we have used only 5 concatenation layers in our method.

Some of qualitative results have been shown in figure 6 to compare small object detection accuracy.

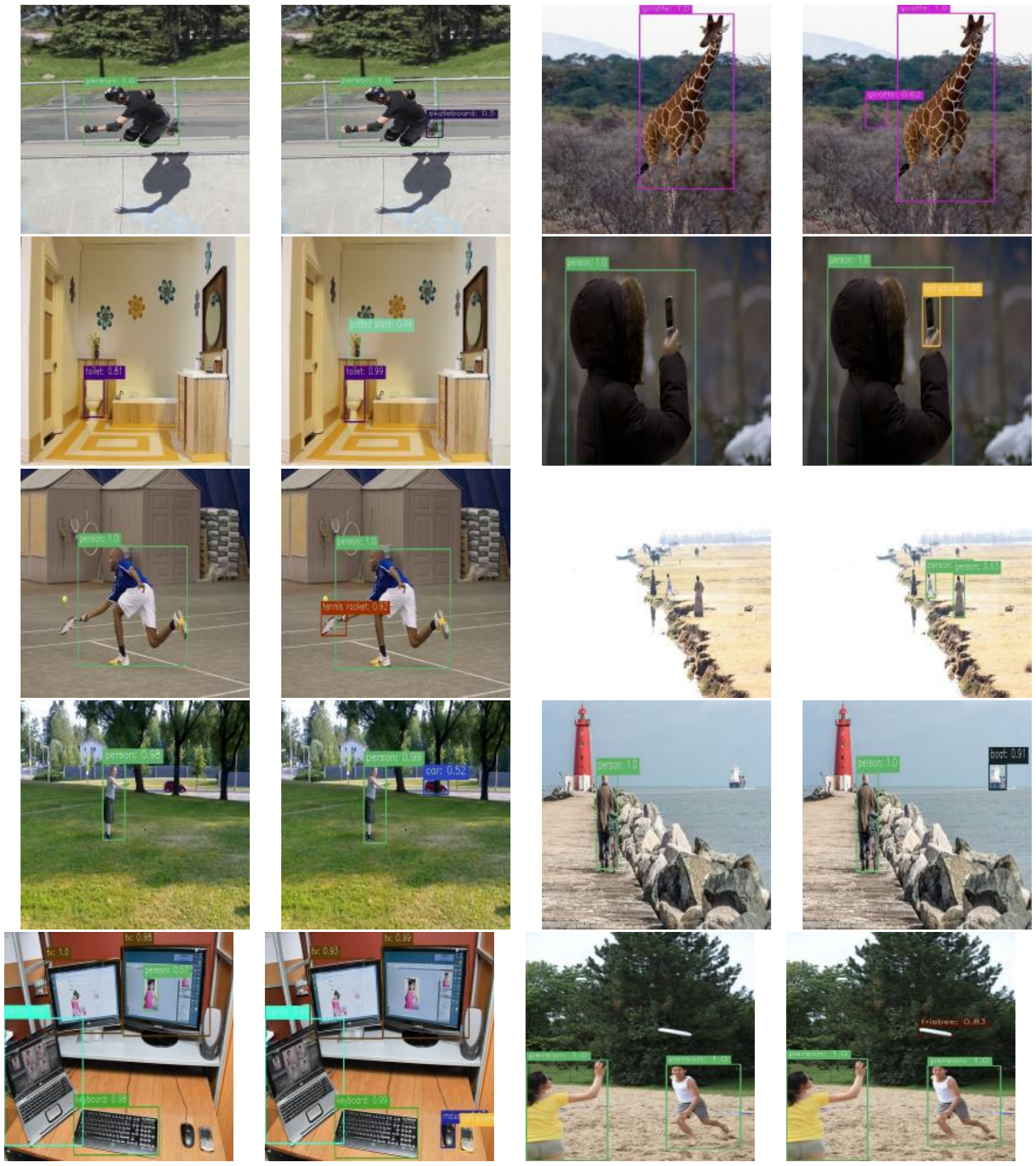

Figure 6: Qualitative results to compare small object detection accuracy. In each pair, left image is mobilenet output and right one is our proposed method.

## 5. Conclusion and Future Work

According to our research, the most important problems of one-stage methods in object detection is the difficulty to detect small objects as well as inaccurate localization. As these issues are combined with hardware limitations, they have become an important challenge. We attempted to tackle hardware limitations while striving to keep the process in the real-time range. Our proposed method increased computational cost slightly. We were able to increase the accuracy of the state-of-the-art system by 0.8% on the COCO dataset, 1.3% on the Pascal VOC and 22.8 fps on Pixel 3 (Snapdragon 845). In our experience on different and smaller datasets, it is still possible to achieve better accuracy by using backbone changes. Also, it is possible to propose better Necks and Heads with less computational costs but no change in accuracy for embedded object detection problem.